\documentclass[cameraready]{Interspeech}

\usepackage[utf8]{inputenc}
\usepackage{ipa}
\usepackage{tipa}
\usepackage{float}
\usepackage{newunicodechar}
\newunicodechar{ᶕ}{\textschwa\textsubscript{l}}
\newunicodechar{ʃ}{\textesh}
\newunicodechar{ʤ}{d\kern-0.22em\textyogh}
\newunicodechar{ʧ}{t\kern-0.18em\textesh}
\newunicodechar{ʒ}{\textyogh}
\newunicodechar{ð}{\eth}
\newunicodechar{ŋ}{\eng}
\newunicodechar{ɛ}{\textepsilon}
\newunicodechar{ɹ}{\textturnr}
\newunicodechar{ɪ}{\textsci}
\newunicodechar{ə}{\textschwa}
\newunicodechar{ʊ}{\textupsilon}
\newunicodechar{ɔ}{\textopeno}
\newunicodechar{ɡ}{g}
\newunicodechar{ɾ}{\textfishhookr}
\newunicodechar{ɑ}{\textscripta}
\newunicodechar{ɒ}{\textturnscripta}
\newunicodechar{ʔ}{\textglotstop}
\newunicodechar{ʌ}{\textturnv}
\newunicodechar{ɚ}{\textschwa\textrhoticity}
\newunicodechar{ɝ}{\textrevepsilon\textrhoticity}
\newunicodechar{ɜ}{\textrevepsilon}
\newunicodechar{θ}{\texttheta}
\newunicodechar{ː}{\textlengthmark}
\newunicodechar{ʴ}{\textsuperscript{\textturnr}}
\newunicodechar{ɣ}{\textgamma}
\newunicodechar{ʰ}{\textsuperscript{h}}

\usepackage{xcolor}

\usepackage{graphicx}
\usepackage{subcaption}

\title{Scaling Human and G2P Supervision for Robust Phonetic Transcription}

\author[affiliation={1}]{Alexander}{Metzger}
\author[affiliation={1}]{Aruna}{Srivastava}
\author[affiliation={1}]{Ruslan}{Mukhamedvaleev}
\address{Koel Labs LLC, USA}
\email{\{alex, aruna, ruslan\}@koellabs.com}


\usepackage{comment}

\begin{document}

\maketitle

\begin{abstract}
    

    Expert phonetic annotation is costly, especially for non-standard dialects and atypical speech. A common alternative is using Grapheme-to-Phoneme (G2P) models to auto-generate phonetic labels from text transcripts at scale. We study how automatic phonetic transcription performance scales with human and G2P supervision in English. Using a curated 80-hour benchmark spanning native, non-native and post-stroke speech, we identify a supervision quality threshold: G2P supervision helps only when fewer than 20-30 hours of human annotation are available. Beyond this threshold, it provides no significant benefit and can reduce cross-dialect robustness. What is effective after this threshold is ASR pretraining which we use to achieve a 2.3× reduction in weighted phone feature error rate over prior systems, with strong gains on non-native and aphasic speech. These results suggest that quantity-driven G2P scaling may yield diminishing returns for robust generalization.
\end{abstract}

\section{Introduction}
Modeling fine-grained phonetic variation is central to many speech technologies, including speech disorder and dementia assessment, pronunciation training, multilingual pretraining, speech enhancement, and voice conversion \cite{CaiSpeechDisorder, DevahiSpeechDementia, shahinPhoneticsForSpeechAssessment, LaneClassicalAustralianLowResource}. While orthographic transcription has become widely accessible through large-scale ASR systems \cite{radford2022whisper}, written text does not capture how sounds are actually realized acoustically. Automated phonetic transcription, typically expressed in the International Phonetic Alphabet (IPA), is therefore critical for applications that depend on detailed acoustic-phonetic information, particularly in clinical and high-stakes settings \cite{PirklbauerPhoneticsForSpeechEnhancement, 
LiuPhoneticsForSpeechEnhancement,
LeePhoneticsForVoiceConversion, ShanPhoneticsForVoiceConversion,
FengPhoneticsForMultilingualPretraining, 
YusuyinPhoneticsForMultilingualPretrainingMCL-ASR,  
GongPhoneticsForSpeechAssessment}.

Progress in phonetic transcription is constrained by the scarcity of expert-annotated data. Producing reliable phonetic labels requires substantial training and inter-annotator agreement \cite{ShribergAnnotatorBiasBroadAndNarrow}. The challenge is amplified for non-native and atypical speech, where annotators must account for cross-linguistic phonological influence and greater acoustic variability \cite{data_epadb}. Yet these speaker populations are precisely those for whom robust phonetic modeling is most important, both for inclusive system design \cite{DavidsonL2PhonologicalDifferencesClassical, MehrabiFairnessML} and for applications like computer-assisted pronunciation training and speech pathology \cite{motivation_americantesol2025englishfluency, motivation_babbel2020accent}.

To address limited human-annotated data, many recent systems rely on Grapheme-to-Phoneme (G2P) models \cite{rao2015grapheme, yolchuyeva19_interspeech} to generate phonetic labels at scale \cite{zero_shot_phone}. By converting orthographic transcripts from large ASR corpora into approximate phoneme sequences, G2P enables training on thousands of hours of machine-labeled data. This approach implicitly assumes that label quantity can compensate for label noise. However, G2P reflects canonical pronunciations associated with a particular dialect and does not capture speaker-specific realizations, mispronunciations, or atypical productions. As a result, large-scale G2P supervision may bias models toward standard pronunciations and reduce sensitivity to acoustic variation \cite{motivation_grapheme_to_phoneme_survey}.

Rather than proposing a new architecture, we study how supervision quality and quantity interact in phonetic transcription. We ask how much expert annotation is needed before large-scale G2P supervision provides diminishing returns. Through a controlled scaling study varying pretraining, ASR, G2P, and human labels, we identify quality thresholds and hypothesize that diverse human annotation is critical to improving robustness to dialectal and atypical speech. Our contributions are fourfold:
\begin{itemize}
\item We curate a standardized 80-hour English benchmark with expert phonetic annotations and diverse speakers enabling controlled evaluation of cross-dialect generalization.
\item We conduct a systematic scaling study of supervision quality and quantity, isolating the effects of self-supervised pretraining, ASR labels, G2P labels, and human labels.
\item We identify a supervision quality threshold effect: G2P supervision improves performance only when fewer than 20-30 hours of diverse human annotation are available.
\item Using this curriculum, we achieve a 2.3× reduction in weighted phone feature error rate over prior systems, with particularly strong gains on non-native and aphasic speech.
\end{itemize}

Because non-native and impaired speakers constitute a substantial portion of English users worldwide \cite{motivation_americantesol2025englishfluency, motivation_babbel2020accent}, we explicitly evaluate across L2 backgrounds and atypical speech conditions. To support reproducibility, we open-source all training code, processing scripts, models, and standardized datasets: \url{https://github.com/KoelLabs/ML}.

\begin{table*}[ht!]
\centering
\captionsetup{labelfont={bf}}
\caption{\textbf{Dataset Statistics and Specifications}. We curate 8 diverse speech corpora for a combined 17 dialects (8 non-native accents) and 1 speech impediment (various severities of post stroke aphasia) across 1,171 speakers and 80 hours of audio processed and split into 41 hours of training data and 14 hours of test data (55 hours in total). We list dialects to the granularity annotated by the datasets. }
\label{tab:dataset_details}
\resizebox{\textwidth}{!}{
\begin{tabular}{lccccccccc}
\toprule
\textbf{Dataset} & \textbf{Speakers} & \textbf{Unproc. (hrs)} & \textbf{Proc. (hrs)} & \textbf{Samples} & \textbf{Dialect/L1} & \textbf{Trans.} & \textbf{Annotators} & \textbf{Test?} \\
\midrule
DARPA TIMIT \cite{data_timit} & 630 & 5.38 & 5.08 & 6300 & 8 US dialects (eus) & Narrow & 3 & \checkmark \\
EpaDB \cite{data_epadb} & 50 & 2.75 & 2.75 & 3160 & Argentinian Spanish (arg) & Narrow & 2 & \checkmark \\
PSST \cite{data_aphasia, data_aphasia_tools} & 107 & 3.58 & 3.58 & 3291 & Post-stroke Aphasia & Broad & 2 & \checkmark \\
L2-ARCTIC (scripted) \cite{data_arctic} & 24 & 3.66 & 3.66 & 3599 & vie, cmn, hin, spa, ara, kor & Broad & 3 & - \\
L2-ARCTIC (spontaneous) \cite{data_arctic} & 22 & 0.44 & 0.39 & 129 & vie, cmn, hin, spa, ara, kor & Broad & 3 & - \\
Speech Ocean \cite{data_speechocean} & 250 & 5.56 & 2.33 & 4960 & Mandarin (cmn) & Broad & 5 & \checkmark \\
Buckeye \cite{data_buckeye} & 40 & 37.96 & 26.45 & 3026 & Columbus, OH (eus) & Narrow & 4 & - \\
DoReCo \cite{data_doreco} & 2 & 2.42 & 0.79 & 577 & Southern British (eng) & Narrow & 2 & - \\
ISLE \cite{data_isle} & 46 & 18.31 & 9.72 & 7714 & Italian (ita), German (deu) & Broad & 6 & \checkmark \\
\bottomrule
\end{tabular}
}
\end{table*}

\section{Methods}

We standardize 80 hours of phonetically annotated English speech spanning diverse dialects and post-stroke aphasia. We systematically compare four curriculum stages: self-supervised pretraining, multilingual ASR finetuning, G2P-generated label finetuning, and human expert label finetuning. Through controlled ablations and scaling experiments, we isolate the effects of label quality, data scale, and curriculum composition on cross-dialect and atypical speech generalization.

\subsection{Datasets}
\label{sec:datasets}

We curate 80.06 hours of English speech across 8 datasets that contain high quality human annotations: TIMIT \cite{data_timit}, L2-ARCTIC \cite{data_arctic} (scripted and spontaneous), EpaDB \cite{data_epadb}, Speech Ocean \cite{data_speechocean}, Buckeye \cite{data_buckeye}, PSST \cite{data_aphasia}, DoReCo \cite{data_doreco}, and ISLE corpus \cite{data_isle}. These span 9 native dialectal variants, L2 speech from 8 L1 backgrounds (Spanish, Korean, Arabic, Vietnamese, Hindi, Mandarin, Italian, and German), and post-stroke aphasia as detailed in Table \ref{tab:dataset_details}. We reserve the ISLE corpus of Italian and German speakers for an unseen dialect test set and use the test sets of TIMIT, EpaDB, PSST, and Speech Ocean for known dialects of unseen speakers. This ensures that our experiments measure generalization, not in-distribution performance.

\subsection{Preprocessing}
\label{sec:data_cleaning}
Phonetic annotation methods vary significantly across datasets, making standardization necessary though non-trivial. Datasets employ ARPAbet, X-SAMPA, TIMITbet, and ESPS phone sets at varying transcription granularities that we carefully map and augment to the correct IPA phones. Beyond symbol mapping, we observe notable quality issues in datasets that we mitigate via non-speech filtering, speaker separation, missing section relabeling, and timestamp correction. After preprocessing, 80.06 hours of raw audio yields 54.75 hours of clean speech (40.81 hours training, 13.94 hours test). The removed audio is mainly silence and non-target speakers, not valid labeled speech since we do not want to bias the speaker distribution. Detailed patch files for our data cleaning are provided in the code.

\subsection{Evaluation Metric}
For evaluation, we use a weighted Phone Feature Error Rate (WPFER) provided by PanPhon \cite{lib_panphon}. A naive Phone Error Rate (PER) treats all phone substitutions equally, so scores are heavily influenced by whether a model's phoneme vocabulary happens to overlap with that of the test set. WPFER addresses this by weighting substitutions according to the number of differing articulatory features between phones, giving partial credit to phonetically similar predictions and providing a fairer reflection of a model's phonetic understanding. The score is normalized by the reference sequence length. We plot 95\% confidence intervals and report statistical significance using 2000 bootstrap samples where each resamples the WPFER distribution \cite{efron1994introduction}.  

\subsection{Hyperparameters}
To ensure observed trends are not side effects of imperfect hyperparameters, we perform full sweeps for each experiment using an 80\% train and 20\% validation split of the non-test data. We base the ranges on prior work \cite{zero_shot_phone, model_vitouphy}: learning rate $\in [3\cdot10^{-5}, 5\cdot10^{-4}]$ (log), warm-up steps $\in [200, 2000]$ (linear), batch size $\in [16, 64]$ (linear), num epochs $\in [3, 20]$ (linear), weight decay $\in [0.00, 0.50]$ (log), time masking $\in [0.05, 0.15]$ (linear), feature masking $\in [0.00, 0.20]$ (linear). All sweeps can be trained in 730 hours using an A100 GPU with 40GB VRAM. 

\begin{table*}[ht!]
\centering
\captionsetup{labelfont={bf}}
\caption{\textbf{WPFER ($\downarrow$) on High Quality Human Annotated Test Sets}. Best WPFER is in bold, second best underlined. For brevity, the top checkpoint of each approach of interest is listed. Full evaluations on 30 prominent models can be found at \url{https://huggingface.co/KoelLabs}. We list both the number of hours of training data on machine derived and human annotated phoneme labels. This excludes non-phoneme training data like XLSR pretraining. *Allosaurus hours are estimated by \cite{model_HuPER}.}
\label{tab:model_eval}
\resizebox{\textwidth}{!}{
\begin{tabular}{lccccccccc}
\toprule
\textbf{Model} & \textbf{Train Machine (hrs)} & \textbf{Train Human (hrs)} & \textbf{TIMIT} & \textbf{EpaDB} & \textbf{PSST} & \textbf{Speech Ocean} & \textbf{ISLE} & \textbf{Avg. WPFER} \\
\midrule
Espeak G2P Baseline \cite{phonemizer_g2p} (2021) & - & - & \underbar{4.97\%} & \underbar{3.9\%} & 25.8\% & \textbf{2.3\%} & \textbf{3.5\%} & 8.1\% \\
\midrule
Allosaurus \cite{allosaurus} (2020) & 2.6K* & 0 & 10.0\% & 13.3\% & 23.9\% & 15.1\% & 10.5\% & 14.6\% \\
W2V2-eSpeak \cite{zero_shot_phone} (2022) & 5.3K & 0 & 5.9\% & 5.9\% & \underbar{13.5\%} & 9.0\% & 4.7\% & \underbar{7.8\%} \\
W2V2-Vitouphy \cite{model_vitouphy} (2023) & 0 & 2.28 & 6.0\% & 7.8\% & 13.8\% & 10.7\% & 5.7\% & 8.1\% \\
MultIPA \cite{model_multiipa} (2023) & 3.6K & 0 & 12.3\% & 10.8\% & 18.8\% & 14.8\% & 8.0\% & 13.0\% \\
HuBERT Phoneme \cite{model_HuBERT_phoneme} (2024) & 0.1K & 0 & 6.4\% & 8.2\% & 20.0\% & 12.8\% & 5.4\% & 10.6\% \\
POWSM \cite{model_powsm} (2025) & 17.1K & 0 & 9.3\% & 7.0\% & 14.4\% & 10.3\% & 4.1\% & 9.0\% \\
WavLM HuPER \cite{model_HuPER} (2026) & 0.1K & 3.7 & 6.7\% & 7.6\% & 14.4\% & 9.8\% & 6.2\% & 9.0\% \\
\midrule
\textbf{Optimal Curriculum (ours)} & 0 & 40.8 & \textbf{2.1\%} & \textbf{2.5\%} & \textbf{5.3\%} & \underbar{3.7\%} & \underbar{3.6\%} & \textbf{3.5\%} \\
\bottomrule
\end{tabular}
}
\end{table*}

\subsection{Curriculum Components}

We analyze four stages of self-supervised and supervised training strategies to optimize model performance. Each stage presents a unique set of limitations in terms of scale and data quality which existing models address distinctly. We perform experiments at each stage to isolate which methods are the most effective in pretraining architecture, balancing data quantity to quality, speaker generalization, and use of curriculum.

\textbf{Stage 1: Large Scale Self-supervised Pretraining.}

Unlabeled speech data are available in many more languages and for much more diverse speakers than labeled data. State of the art speech models take advantage of this through various self-supervised pretraining strategies \cite{model_xlsr, model_hubert, model_wavlm}.

Wav2Vec2 XLSR \cite{model_xlsr} uses self-supervised contrastive learning directly on raw waveform audio from 60K hours of unlabeled multilingual speech, yielding strong low-resource performance, cross-lingual transfer, and phoneme-level modeling. HuBERT \cite{model_hubert} clusters acoustic features to learn hidden units with existing checkpoints pretrained on 60K hours of speech. WavLM \cite{model_wavlm} builds on this with an extra denoising objective by simulating overlapped speech on 94K hours, making these pretrained checkpoints more robust to noise and achieving SoTA performance on advanced speech tasks like speaker separation.

Section \ref{sec:prior_eval} evaluates how these approaches perform for phone transcription to select the foundation for our experiments.

\textbf{Stage 2: Multilingual Supervised ASR finetuning.}

Beyond unlabeled pretraining, Automatic Speech Recognition (ASR) is a well-studied task with high quality text transcription datasets abundant compared to scarce phonetic annotations. Xu et al. \cite{zero_shot_phone} demonstrated that a checkpoint pretrained on 53 languages \cite{model_xlsr53} can yield lower error rates for phone transcription when fine-tuned on machine labeled phonetic data while also performing well on ASR tasks. 

We evaluate the impact of combining ASR models with phonetic finetuning when scaling data quality in Section \ref{sec:curriculum_ablations}.

\textbf{Stage 3: Machine Generated Phone Labels.} 

A straightforward approach to leveraging abundant ASR data for phone transcription is to approximate the phonetic labels from the orthographic text labels using a Grapheme-to-Phoneme (G2P) model. This approach is adopted by the vast majority of models \cite{zero_shot_phone, model_ginic, model_HuBERT_phoneme, model_HuPER, model_multiipa, model_powsm}. However, it naturally does not model mispronunciations, accents, or other characteristics not captured in orthographic transcriptions. 

This observation motivates our investigation of how well G2P-derived labels scale to non-native dialects and speech impediments in Section \ref{sec:curriculum_ablations} and \ref{sec:human_and_g2p_scaling}.

\textbf{Stage 4: Human Annotated Phone Labels.} 
Human phonetician annotations are generally considered the gold standard for phone transcription \cite{model_HuPER}, as they directly reflect produced speech. However, they require deep expertise and substantial annotation time, limiting availability for the majority of languages and dialects. Several models have explored combining human with machine-labeled data at the scale of 5 or fewer hours of human data, both as a preconditioning step \cite{model_HuPER} and as subsequent fine-tuning \cite{model_vitouphy, model_mrrubino}. However, per Section \ref{sec:prior_eval}, these have not outperformed purely machine-labeled data at scale.

This motivates us to ask whether combining human and machine-labeled data could be successful when the human data are scaled. In Section \ref{sec:human_and_g2p_scaling}, we use our curated 40 hours of training data to analyze how well the best model architecture and pretraining setup performs with 0\%, 25\%, 50\%, 75\%, and 100\% of the human-labeled fine-tuning data. To isolate the effect of label quality, we keep the audio data constant and swap out proportions of human labels with G2P labels. We additionally conduct experiments with and without 5.3K hours of G2P labels \cite{zero_shot_phone} to reflect the practical advantage G2P: scalability.

\section{Results}
Our optimal curriculum—XLSR pretraining, ASR fine-tuning, and 40 hours of human phonetic supervision—achieves a 2.3× WPFER reduction over prior systems and substantially improves on non-native and aphasic speech. Ablations show that ASR pretraining improves out-of-domain generalization, while large-scale G2P augmentation provides little benefit and can degrade performance once 20–30 hours of human-labeled data are available. These results demonstrate that moderate amounts of high-quality human annotation feasibly outperform quantity-driven machine labeling for robust phone transcription.

\subsection{Comparison to Prior Approaches}
\label{sec:prior_eval}
We compare the 30 most prominent phone transcription models in literature from 2020 to 2026 and summarize the top scores for 7 approaches of interest in Table \ref{tab:model_eval}. Notably, all perform poorly for PSST \cite{data_aphasia_tools} which contains impaired speech from speakers suffering from post-stroke aphasia. These samples are often not coherent words or contain significant stuttering meaning models cannot rely on memorizing what phonemes typically comprise a word and instead must base prediction on the true pronunciation. Of the existing self-supervised approaches and architectures, Wav2Vec2 \cite{model_wav2vec2} with XLSR \cite{model_xlsr} has the lowest error rates. 
As a baseline, we evaluate which G2P label method most closely matches the human labels and find that Espeak \cite{espeak_speech_synthesizer} labels slightly outperform Phonemizer \cite{phonemizer_g2p} and CMUDict \cite{cmudict_g2p} (this holds when speech models are trained using their respective labels as well). Note that this isn't a direct comparison to the speech models since we run G2P on the ground-truth ASR transcripts, but it does show that G2P performs significantly better for broadly annotated datasets like Speech Ocean and ISLE and significantly worse for impaired speech from PSST which helps explain why most models trained with G2P labels significantly underperform on PSST. Motivated by this, we investigate when including G2P data may negatively affect performance especially when human data are available in Sections \ref{sec:curriculum_ablations} and \ref{sec:human_and_g2p_scaling}.
\begin{figure}[H]
  \centering
  \includegraphics[width=\linewidth]{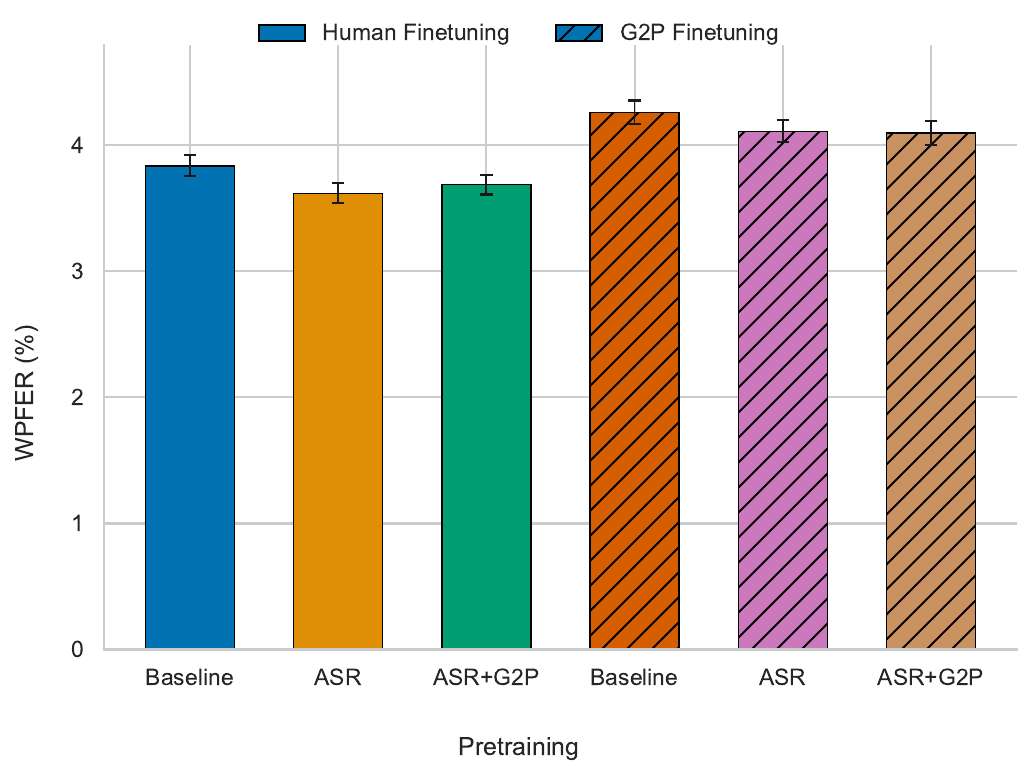}
  \caption{\textbf{Out of Domain WPFER ($\downarrow$) by Curriculum} using unseen German and Italian speakers (ISLE \cite{data_isle}). Finetuning with human data significantly outperforms finetuning with G2P labels. Regardless of finetuning data, extending the XLSR baseline with additional ASR pretraining yields significantly lower error rates ($p < 0.05$). ASR pretraining alone matches or exceeds 5.3K hours of additional G2P pretraining. }
  \label{fig:curriculum_ablation}
\end{figure}

\begin{figure*}[ht!]
  \centering
  \includegraphics[width=\linewidth]{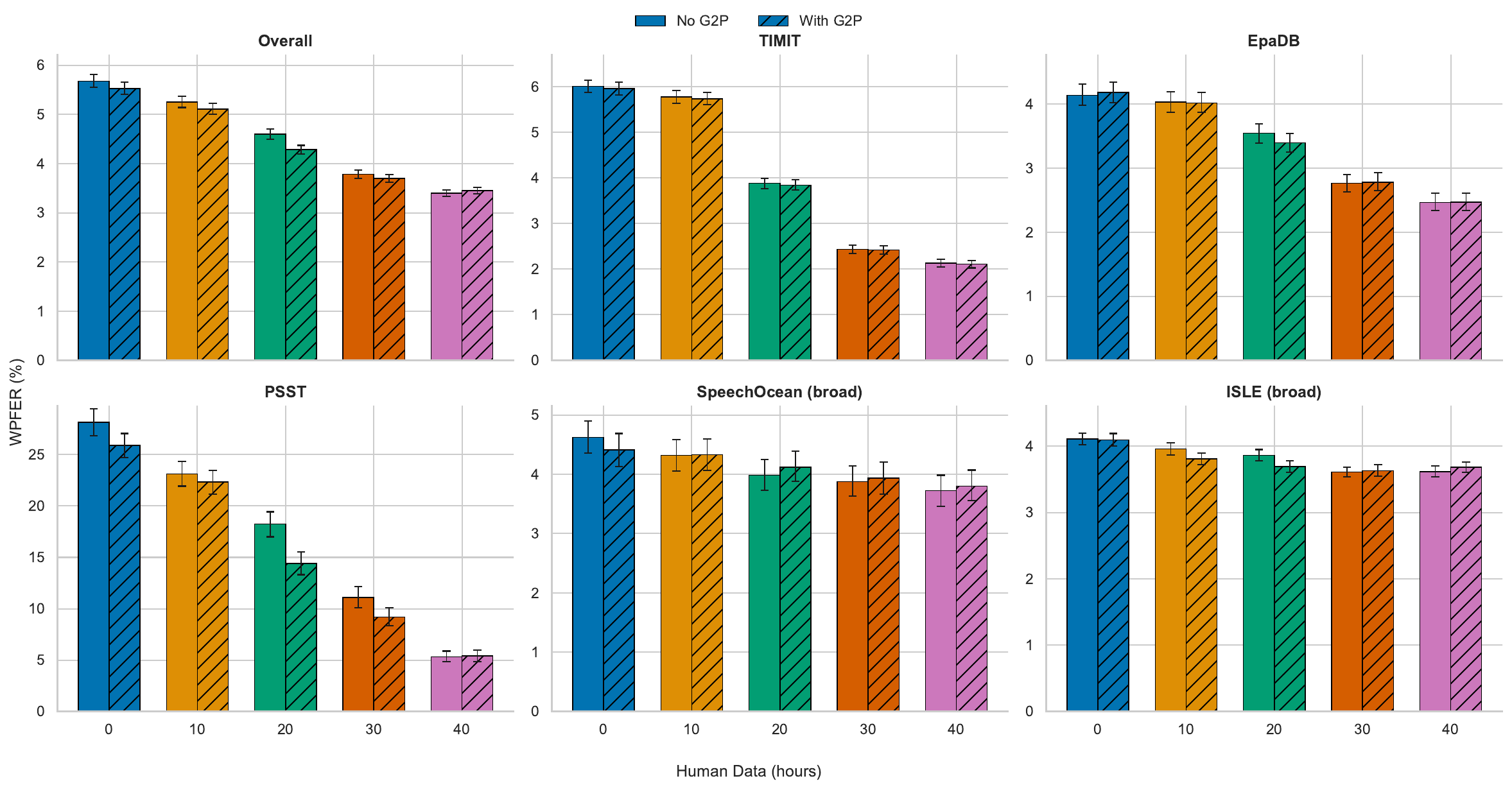}
  \caption{\textbf{WPFER ($\downarrow$) Scaling with Human and G2P Data.} As expected, increasing the amount of human data significantly improves performance ($p < 0.05$) both with no G2P pretraining and with G2P pretraining. The effect is larger for narrowly transcribed datasets like TIMIT, EpaDB, and PSST, where an increment of 10 hours or less is enough to significantly reduce WPFER ($p < 0.05$) and less so for broadly transcribed datasets like Speech Ocean and ISLE where at least 20 hours of data are needed for a significant reduction. Interestingly, scaling up the amount of G2P pretraining data only helps when there is less than 20-30 hours of human training data.}
  \label{fig:data_quality_ablation}
\end{figure*}

\subsection{Curriculum Component Ablations}
\label{sec:curriculum_ablations}

Figure \ref{fig:curriculum_ablation} illustrates WPFER on unseen German and Italian accents \cite{data_isle} with different curricula of pretraining and finetuning data. The lowest error rate is achieved when G2P is excluded and the curriculum is comprised of ASR pretraining + 40 hours of human-annotated data. Extending the curriculum with G2P data contributes no statistically significant difference and even slightly degrades generalization to the ISLE dialects when finetuning on human data. ASR pretraining on the other hand does significantly lower error rates regardless of finetuning data ($p < 0.05$). This is likely because the ASR intermediary introduces the model to greater speaker and dialect variety without biasing the model towards G2P's adherence to standard speech patterns. Similarly, finetuning on human data also significantly lowers error rates ($p < 0.05$), even when it is less in quantity than G2P data by orders of magnitude. This shows that high quality labels that match true pronunciation are critical to generalize to unseen dialects and that this cannot be achieved by scaling machine labels alone. Previous systems \cite{zero_shot_phone, model_powsm, model_HuBERT_phoneme} rely heavily on G2P labels, with no prior prominent system reporting more than 5 hours of human-annotated training data. To the contrary, our results suggest that greater focus should be placed on the more robust ASR pretraining and obtaining higher quality labels. The natural next question is how much high quality data is needed and how ``high quality'' the data needs to be which is what Section \ref{sec:human_and_g2p_scaling} will analyze. 

\subsection{Scaling Trends of Human and G2P Data}\label{sec:human_and_g2p_scaling}
As expected, Figure \ref{fig:data_quality_ablation} shows progressive improvement as the proportion of human-labeled data increases relative to G2P-labeled data. The full human-labeled dataset achieves a mean WPFER of 3.4\% (95\% CI [3.3, 3.5]), whereas the full G2P-labeled dataset scores a mean WPFER of 5.5\% (95\% CI [5.4, 5.7]). For narrowly transcribed datasets such as PSST \cite{data_aphasia_tools}, EpaDB \cite{data_epadb}, and TIMIT \cite{data_timit}, an increment of 10 hours of human-labeled data is sufficient to significantly reduce error rates by up to 5\% WPFER ($p < 0.05$). Broadly transcribed datasets such as Speech Ocean \cite{data_speechocean} and ISLE \cite{data_isle} require at least 20 hours of human-labeled data to achieve a significant reduction. This suggests that for more nuanced applications like pronunciation feedback and clinical diagnosis, high quality human finetuning data is essential, whereas for more coarse applications the gain is minimal. G2P pretraining on 5.3K hours of speech proves beneficial when human-labeled data is scarce (less than 20-30 hours), reducing error rates by up to 4\% WPFER on PSST \cite{data_aphasia_tools}, consistent with prior findings \cite{zero_shot_phone}. However, once human-labeled data exceeds this threshold, the inclusion of G2P data yields no significant reduction in error rate and frequently degrades performance. While the scale afforded by G2P labeling remains valuable, it is less critical than previously hypothesized, suggesting that collecting enough human data might be feasible in many use cases. Given machine-generated labels are less effective when combined with medium-scale human-labeled data, future training strategies should be evaluated on larger human-annotated datasets to assess scalability.

\section{Conclusion}
We systematically evaluated how phonetic transcription performance scales with human and G2P supervision. In English, we find that once approximately 20–30 hours of diverse human-annotated speech are available, additional G2P supervision provides no statistically significant benefit and can slightly degrade on cross-dialect and atypical-speech. In contrast, ASR pretraining consistently improves out-of-domain generalization without introducing label bias. These findings suggest that moderate amounts of high-quality phonetic annotation may be more effective than quantity-driven machine labeling.

Our study is limited to English and to the curated dialects considered. Future work should evaluate whether similar scaling thresholds hold across languages with different phonemic inventories and under more extreme low-resource conditions.

\ifcameraready
\section{Acknowledgments}
We thank Mozilla and Google for funding the training runs and the annotators and dataset maintainers for their guidance in consolidating notation schemes and correcting annotation mistakes. Also huge thanks to Eunjung for being amazing.
\fi

\section{Generative AI Use Disclosure}
The authors used generative AI for editorial purposes to correct grammar and spelling mistakes while preparing this manuscript.

\bibliographystyle{IEEEtran}
\bibliography{mybib}

\end{document}